\documentclass[conference]{IEEEtran}
\IEEEoverridecommandlockouts

\usepackage{cite}
\usepackage{amsmath,amssymb,amsfonts}
\usepackage{algorithmic}
\usepackage{graphicx}
\usepackage{textcomp}
\usepackage{xcolor}

\usepackage{caption}
\usepackage{subcaption}
\usepackage{algorithm,algorithmic}
\usepackage{booktabs,multirow,siunitx}

\ifodd 1
\definecolor{darkgreen}{RGB}{0,200,0}
\else

\fi

\def\BibTeX{{\rm B\kern-.05em{\sc i\kern-.025em b}\kern-.08em
    T\kern-.1667em\lower.7ex\hbox{E}\kern-.125emX}}
\begin{document}

\title{LCFed: An Efficient Clustered Federated Learning Framework for Heterogeneous Data
}

\author{\IEEEauthorblockN{1\textsuperscript{st} Yuxin Zhang}
\IEEEauthorblockA{\textit{School of Computer Science} \\
\textit{Fudan University}\\
Shanghai, China \\
yxzhang24@m.fudan.edu.cn}
\\
\IEEEauthorblockN{4\textsuperscript{th} Zhe Chen}
\IEEEauthorblockA{\textit{School of Computer Science} \\
\textit{Fudan University}\\
Shanghai, China \\
zhechen@fudan.edu.cn}
\and
\IEEEauthorblockN{2\textsuperscript{nd} Haoyu Chen}
\IEEEauthorblockA{\textit{School of Computer Science} \\
\textit{Fudan University}\\
Shanghai, China \\
21110240001@m.fudan.edu.cn}
\\
\IEEEauthorblockN{5\textsuperscript{th} Jin Zhao$^*$}
\IEEEauthorblockA{\textit{School of Computer Science} \\
\textit{Fudan University}\\
Shanghai, China \\
jzhao@fudan.edu.cn}
\and
\IEEEauthorblockN{3\textsuperscript{rd} Zheng Lin}
\IEEEauthorblockA{\textit{Department of Electrical and} \\
\textit{Electronic Engineering} \\
\textit{The University of Hong Kong}\\
Hong Kong, China \\
linzheng@eee.hku.hk}\\

\thanks{Copyright 2025 IEEE. Published in ICASSP 2025 – 2025 IEEE International Conference on Acoustics, Speech and Signal Processing (ICASSP), scheduled for 6-11 April 2025 in Hyderabad, India. Personal use of this material is permitted. However, permission to reprint/republish this material for advertising or promotional purposes or for creating new collective works for resale or redistribution to servers or lists, or to reuse any copyrighted component of this work in other works, must be obtained from the IEEE. Contact: Manager, Copyrights and Permissions / IEEE Service Center / 445 Hoes Lane / P.O. Box 1331 / Piscataway, NJ 08855-1331, USA. Telephone: + Intl. 908-562-3966.}
}

\maketitle

\begin{abstract}
Clustered federated learning (CFL) addresses the performance challenges posed by data heterogeneity in federated learning (FL) by organizing edge devices with similar data distributions into clusters, enabling collaborative model training tailored to each group.
However, existing CFL approaches strictly limit knowledge sharing to within clusters, lacking the integration of global knowledge with intra-cluster training, which leads to suboptimal performance. Moreover, traditional clustering methods incur significant computational overhead, especially as the number of edge devices increases.
In this paper, we propose LCFed, an efficient CFL framework to combat these challenges.
By leveraging model partitioning and adopting distinct aggregation strategies for each sub-model, LCFed effectively incorporates global knowledge into intra-cluster co-training, achieving optimal training performance.
Additionally, LCFed customizes a computationally efficient model similarity measurement method based on low-rank models, enabling real-time cluster updates with minimal computational overhead.
Extensive experiments show that LCFed outperforms state-of-the-art benchmarks in both test accuracy and clustering computational efficiency.
\end{abstract}


\section{Introduction}
Deep learning (DL) is rapidly advancing, fueled by unprecedented amounts of data and achieving significant successes in areas such as signal processing~\cite{liu2024audioldm,yuan2024satsense,peng2024sums,lin2022channel,yuan2023graph,zhao2024leo}, anomaly detection~\cite{vos2024privacy}, smart healthcare~\cite{zou2022variational,fang2024ic3m,tang2024merit}, and autonomous driving~\cite{hu2024agentscomerge,zheng2023autofed,hu2024agentscodriver}. However, this data is typically distributed across large-scale mobile devices and the Internet of Things (IoT)~\cite{wang2018edge,hu2024accelerating}, with privacy restrictions~\cite{dataprivacy,lin2024efficient} and network bandwidth limitations~\cite{rodio2023federated,lin2024adaptsfl} preventing its transfer to a central server for training. To address these challenges, federated learning (FL)~\cite{FedAvg,lin2024hierarchical,zhang2024fedac,sanchez2024federated,chen2024gradient,lin2024fedsn,zhang2024satfed,zheng2023autofed} has emerged as a privacy-preserving distributed DL paradigm. FL training involves exchanging only model parameters, not raw data, between edge devices and a central server, thereby ensuring privacy and significantly reducing communication overhead~\cite{lin2024split}.


However, data heterogeneity severely impedes the efficiency of democratizing FL across edge devices~\cite{yuan2022convergence,lin2024splitlora,heter1,fang2024automated,lin2023pushing}. The non-independent and identically distributed (non-IID) nature of device raw data often results in disparate local risks, complicating the task of optimizing a single global model for all devices~\cite{leo2025split}. To overcome this limitation, clustered federated learning (CFL)~\cite{CFL,FeSEM} groups edge devices into multiple clusters based on similarities in local data distribution. This approach enables intra-cluster collaborative training of a central model tailored to the specific data characteristics of each cluster, thereby improving overall training performance.



Despite extensive research~\cite{IFCA,CGPFL,zw2024specbreathing}, current CFL methods still face significant limitations.
{Firstly, despite differences in data distribution between clusters, some knowledge remains beneficial for sharing. Existing CFL methods isolate clusters, limiting valuable inter-cluster knowledge exchange and resulting in suboptimal test accuracy.}
Secondly, clustering in CFL necessitates computing the similarity between device models and cluster centers. However, with the rapid expansion of the number of edge devices participating in FL~\cite{wang2018edge}, the computational cost of these similarity measurements has increased substantially, thus hindering training convergence.

To address these challenges, we propose \textit{LCFed}, a novel and efficient CFL framework. LCFed splits the model and employs specialized training and aggregation strategies for distinct sub-models, thereby effectively integrating global knowledge within each cluster to optimize training performance. {Furthermore, LCFed customizes a computationally efficient measure for model similarity. By offline sampling devices to compute a model rank decomposition matrice, LCFed enables the server to update similarities using low-rank models, significantly reducing server-side computational overhead.} Our key contributions are summarized as follows.



\begin{figure*}[t]
\centering
\includegraphics[width=.9\textwidth]{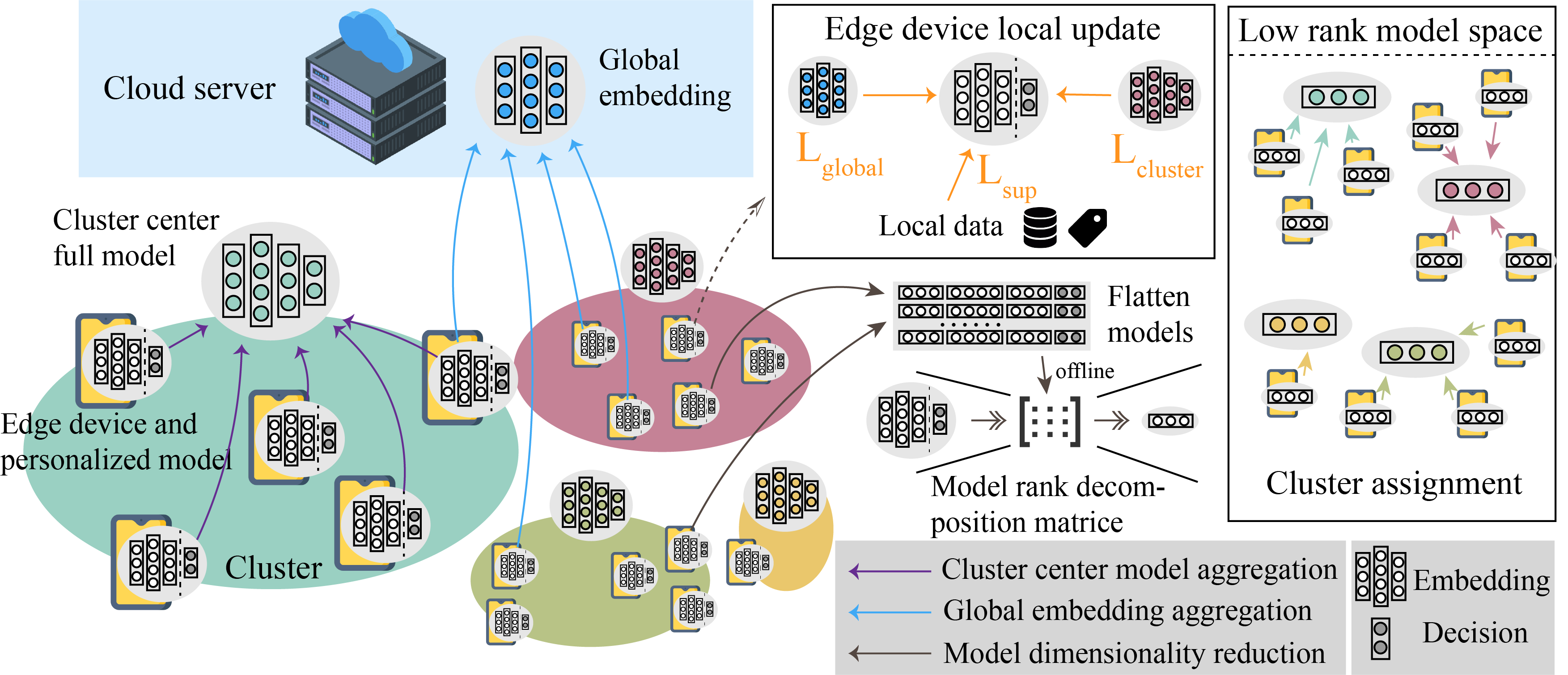}
\caption{The LCFed {framwork} in heterogeneous settings.
{The server clusters and aggregates clients to get a global embedding and multiple cluster center models. Clients update local models by minimizing the classification error loss ($L_{\text{sup}}$), intra-clustering regularization ($L_{\text{cluster}}$), and global regularization ($L_{\text{global}}$).}
}
\label{fig:framwork}
\vspace{-2ex}
\end{figure*}

\begin{itemize}
    \item
In this paper, we propose LCFed, an efficient CFL framework for heterogeneous data. {To the best of our knowledge, LCFed is the first CFL framework to achieve optimal training performance by employing model splitting to facilitate simultaneous knowledge sharing at both the intra-cluster and global levels.}
    \item
{We investigate the computational overhead of clustering in CFL and customize a low-rank model-based similarity measurement method for LCFed. This approach significantly reduces the server’s computational burden for cluster assignments, particularly in large-scale scenarios.}
    \item
We conduct experiments on diverse real datasets in heterogeneous settings to demonstrate the superior overall performance, robustness, and computational efficiency of LCFed, surpassing state-of-the-art methods.
\end{itemize}

\section{Problem Formulation}
\label{sec:formulation}
We consider a typical FL scenario where each device $i \in [m]$ possesses its private dataset $\mathcal{D}_i$ from distribution $ \mathbb{P}_i({\bf{x}}, y)$, where $m$ is the number of devices, $\bf{x}$ and $y$ represent the input features and corresponding labels.
In LCFed, each device $i$ trains a personalized model $ f_i(\omega_i;\cdot) $ with weights $\omega_i$.
The server groups devices into $K$ clusters and trains the cluster center models $\{\Omega_k\}_{k \in [K]}$ to guide the personalized model training within each cluster.
Thus, the objective function is:
\begin{equation}
\begin{split}
&\min_{\{ \omega_i \}} \frac{1}{m} \sum_{i=1}^m  \{ \mathcal{L}_i(\omega_i) + \frac{\mu}{2} ||\omega_i - \Omega_{k^*(i,R^*)}||^2_2 \} ,\\
\textrm{s.t.} \quad &R^* = \underset{R \in \mathbb{R}^{m \ast K}}{\arg\max} \sum_{k=1}^K \sum_{i=1}^m R_{i,k} \textrm{Simi} (\omega_i, \Omega_k(\{ \omega_i \}_{i \in [m]},R)),
\end{split}
\end{equation}
where $\mu$ is a regularization factor, $\textrm{Simi}(\cdot,\cdot)$ measures model similarity, and $R$ is the assignment matrix where $R_{i,k} = 1$ if device $i$ belongs to cluster $k$ else $R_{i,k} = 0$.
$\Omega_k(\{ \omega_i \}_{i \in [m]},R) = \frac{1}{\sum_{i=1}^m R_{i,k}}\sum_{i=1}^m R_{i,k} \omega_i$ is the central model of cluster $k$, and $k^*(i,R)$ represents the cluster to which device $i$ belongs, corresponding to the sole $1$ in the $i$-th row of $R$.
Note that $k^*$ is well-defined per $R$'s definition.

\section{Methodology}
\label{sec:methodology}

The core design of the LCFed framework centers on two key innovations: (1) integrating global knowledge into intra-cluster co-training by employing model partitioning, and (2) minimizing the server-side computational costs of similarity updates through the use of low-rank models. An overview of the LCFed framework is illustrated in Fig~\ref{fig:framwork}.




\subsection{Global and Intra-Cluster Knowledge Integration}
\label{subsec:intergrate_know}

An inherent limitation of existing CFL methods is their confinement of all knowledge sharing within each cluster. This constraint prevents devices from accessing valuable global-scale knowledge beyond their cluster boundaries, thereby leading to suboptimal performance. To overcome this limitation,
we draw inspiration from insights into latent space features of DL models~\cite{FedRep,FedProto}, and split the model $\omega$ into a shallow \textit{embedding} sub-model $\phi$ and a deep \textit{decision} sub-model $h$.
The embedding $\phi$ processes low-dimensional features of input data, providing a foundational representation for subsequent tasks~\cite{FedProto}.
{This representation is less affected by data heterogeneity, making it well-suited for collaborative learning across the global scope~\cite{shareproto1}.}
Meanwhile, the decision $h$ captures device-specific data distributions and is ideally shared within the intra-cluster context~\cite{nosharedeep}.


Specifically, in each training round, the server aggregates device local models $\{ \omega_i = (\phi_i, h_i) \}_{i \in [m]} $ into a global embedding $\Phi = \sum_{i=1}^m  \phi_i$ and $K$ cluster center models $\{ \Omega_k = \frac{1}{\sum_{i=1}^m R_{i,k}}\sum_{i=1}^m R_{i,k} \omega_i \}_{k \in [K]}$.
During local updates, device $i$ learns global knowledge from $\Phi$ and intra-cluster knowledge from its cluster center model $\Omega_{k^*}$.
The optimization objective for local updates is as follows:
\begin{equation}
\begin{split}
\min_{\{ \omega_i \}} \frac{1}{m}\sum_{i=1}^m  \{ \underbrace{\mathcal{L}_i(\omega_i)}_{L_{\text{sup}}} + \frac{\mu}{2}\underbrace{|| \omega_i - \Omega_{k^*} ||^2_2}_{L_{\text{cluster}}} + \frac{\lambda}{2}\underbrace{ || \phi_i - \Phi||^2_2}_{L_{\text{global}}}\} ,
\end{split}
\end{equation}
where $\mu$ and $\lambda$ are regularization factors.

\begin{algorithm}[tb]
\caption{LCFed}
\label{alg:LCFed}
\textbf{Input}: $K$, $\eta$, $\mu$ and $\lambda$\\
\textbf{Output}: $ \{ \omega_i \}_{i \in [m]} $\\
\textbf{Server executes}:
\begin{algorithmic}[1] 
\STATE Initialize: $R^0$, $ \{ \omega_i^0 = (\phi_i^0, h_i^0)\}_{i \in [m]} $,  $ \{ \Omega_k^0 \}_{k \in [K]} $, $\Phi^{0}$
\STATE Compute $M$ based on Sec.~\ref{subsec:lrcos}
\FOR{each round $t = 0,1,...$}
\STATE Randomly selects a subset of devices $S_t$
\FOR{each device $i$ $\in$ $S_t$ \textbf{in parallel}}
\STATE Server sends  $\Omega^{t}_{k^*}$, $\Phi^{t}$ to device $i$
\STATE $ \omega_i^{t+1} \gets $ LocalUpdate ($ \omega_i^{t} $, $ \Omega^{t}_{k^*} $, $\Phi^{t}$, $\mu$, $\lambda$, $\eta$)
\STATE Devices $i$ sends $ \omega_i^{t+1} $,  $ M \cdot \omega_i^{t+1} $ back
\ENDFOR
\STATE Update $R^{t+1}$ based on Eq.~\ref{eq::estep} and Eq.~\ref{eq::lrcos}
\STATE Update $ \Phi^{t+1}$ and $ \{ \Omega_k^{t+1} \}_{k \in [K]}$ based on Sec.~\ref{subsec:intergrate_know}
\ENDFOR
\end{algorithmic}
\textbf{}

\textbf{LocalUpdate ($ \omega_i^{t}$, $\Omega^{t}_{k^*} $, $\Phi^{t}$, $\mu$, $\lambda$, $\eta$)}:

\begin{algorithmic}[1]
\STATE $\omega_i^{t,0} = \omega_i^{t}$
\FOR{each local epoch $r$ = 0,1,...}
\STATE Randomly selects a batch $\mathcal{B}_i$ from $\mathcal{D}_i$
\STATE  $\omega_i^{t,r+1} = \omega_i^{t,r} - \eta\nabla l_i(\omega_i^{t,r};\mathcal{B}_i) -\eta\mu(\omega_i^{t,r} - \Omega^{t}_{k^*}) - \eta\lambda(\phi_i^{t,r} - \Phi^t)$
\ENDFOR
\STATE \textbf{return} $\omega_i^{t,r+1}$
\end{algorithmic}
\end{algorithm}

In LCFed, local training leverages a meticulously designed loss function to capture knowledge across three distinct dimensions: local training data ($L_{\text{sup}}$), global knowledge ($L_{\text{global}}$), and intra-cluster knowledge ($L_{\text{cluster}}$). By balancing these knowledge dimensions through hyperparameters $\mu$ and $\lambda$, LCFed achieves optimal performance, as substantiated in the experimental section.


\subsection{Low-Rank Model Similarity}
\label{subsec:lrcos}

Another challenge for CFL lies in achieving effective and computationally efficient cluster assignment.
Some methods perform clustering only once at the start of training (\textit{offline})~\cite{FLHC,FedGroup}, making the clustering results highly susceptible to early-stage randomness, which compromises robustness and ultimately degrades collaborative training performance.
In contrast, \textit{online} clustering methods continuously update model similarities and cluster assignments at each global round~\cite{FeSEM,IFCA}. While this approach yields more effective and robust clustering outcomes, it also imposes a substantially higher computational burden~\cite{wang2017towards,wang2023accelerate}.
Current CFL methods typically compute model similarity $\textrm{Simi}(\cdot,\cdot)$ using cosine similarity or negative L2 distance between device and cluster center parameters for updating assignment matrix $R$:
\begin{equation}
R_{i,k} = \begin{cases}
1, & k = \underset{j}{\arg\max} \textrm{Simi}( \omega_i , \Omega_j ) \\
0,& \textrm{else} \\
\end{cases}
.
\label{eq::estep}
\end{equation}
Consequently, each training round requires $m \times K$ server-side model similarity measures. As the number of devices \(m\) involved in FL training continue to surge~\cite{wang2018edge,lin2024efficient}, the clustering computational cost becomes increasingly substantial.



To tackle this challenge, LCFed customizes a low-rank model-based similarity measure. Specifically, at the beginning of LCFed training, we offline sample a subset of devices $S_d$, flattening and concatenating their local updated models into $W_d \in \mathbb{R}^{|S_d| \times dim(\omega)}$. We then apply dimensionality reduction techniques (e.g., PCA~\cite{PCA}) to $W_d$, yielding a model rank decomposition matrice $M \in \mathbb{R}^{dim(\omega) \times D}$ with low-rank model space dimensionality $D$. Each device $i$ stores $M$ locally and sends the low rank model $M \cdot \omega_i$, along with its updated model $\omega_i$, to the server in each round. Consequently, the server updates $\textrm{Simi} (\omega_i, \Omega_k)$ based on these low-rank models:
\begin{equation}
\textrm{Simi} (\omega_i, \Omega_k) = \frac{M \cdot \omega_i \cdot M \cdot \Omega_k}{||M \cdot \omega_i||_2||M \cdot \Omega_k||_2}.
\label{eq::lrcos}
\end{equation}
By using low-rank models, LCFed reduces the similarity computation cost by a factor of around $ \frac{D}{dim(\omega)} $, and we will discuss the clustering overhead in detail in subsequent experiments.
The complete pseudocode of the LCFed framework is presented in Algorithm~\ref{alg:LCFed}.

\section{Evalutaion}
\label{sec:evaluation}

\subsection{Experimental Setup}

We implement LCFed framework using Python 3.7 and PyTorch 1.12.1., and train it with NVIDIA GeForce RTX 3090 GPUs.
We adopt the well-known LeNet-5~\cite{MNIST} model, trained with the MNIST~\cite{MNIST}, CIFAR-10 and CIFAR-100~\cite{CIFAR10} datasets.
We split LeNet-5 into the final fully connected layer as decision $h$ and other shallow layers as embedding $\phi$.
Consistent with previous research~\cite{chen2022towards,CGPFL}, we employ the Dirichlet distribution with $\alpha=0.1$ and the pathological distribution (where $n$ denotes the number of labels per device) for non-IID data settings.
A total of $m=100$ devices are simulated, with the number of clusters $K$ set to 10.
The SGD optimizer is employed with a learning rate set to 0.01.
Other hyperparameters include a batch size of 32 and the number of local iterations set to 5.
To investigate the advantages of LCFed, we compare it with six other state-of-the-art benchmarks:
(1) FedAvg~\cite{FedAvg}: all devices share a global model;
(2) FedPer~\cite{FedPer}: devices share a global embedding $\phi$ while retaining personalized $h$;
(3) FedGroup~\cite{FedGroup}: offline clustering based on model cosine similarity;
(4) FeSEM~\cite{FeSEM}: online clustering based on model L2 distances;
(5) CGPFL~\cite{CGPFL}: personalized model training guided by each cluster center;
(6) IFCA~\cite{IFCA}: online clustering based on the empirical risk of each cluster center model on the device.

\subsection{Overall Performance}

\begin{table}[t!]
\centering
    \caption{The LCFed's performance is evaluated against the benchmarks using mean and standard deviation of test accuracy. The best mean accuracy is marked in bold.}
  \begin{tabular}{lSSS}
    \toprule
    \multirow{2}{*}{\textbf{Method}} &
    \multicolumn{3}{c}{MNIST Top-1 Test Accuracy (\%)} \\
    & {$\alpha=0.1$} & {$n=3$} & {$n=4$} \\
      \midrule
    {FedAvg} & {97.98$\pm$0.16} & {97.11$\pm$0.45} & {96.95$\pm$0.43} \\ 
    {FedPer} & {94.86$\pm$0.22} & {96.52$\pm$0.65} & {96.08$\pm$0.59} \\
    {FeSEM} & {96.12$\pm$0.31} & {97.77$\pm$0.28} & {97.13$\pm$0.39} \\
    {FedGroup} & {96.13$\pm$0.35} & {98.08$\pm$0.36} & {97.56$\pm$0.33} \\
    {CGPFL} & {95.15$\pm$0.36} & {97.84$\pm$0.29} & {96.98$\pm$0.61} \\
    {IFCA} & {\textbf{98.41$\pm$0.20}} & {98.35$\pm$0.43} & {97.45$\pm$0.33} \\
    {\textbf{LCFed}} & {97.29$\pm$0.30} & {\textbf{98.53$\pm$0.26}} & {\textbf{98.31$\pm$0.43}} \\
    \bottomrule
    \toprule
    \multirow{2}{*}{\textbf{Method}} &
    \multicolumn{3}{c}{CIFAR10 Top-1 Test Set Accuracy (\%)} \\
    & {$\alpha=0.1$} & {$n=3$} & {$n=4$} \\
      \midrule
    {FedAvg} & {64.75$\pm$1.21}  & {62.73$\pm$0.42} & {61.79$\pm$0.85} \\
    {FedPer} & {70.84$\pm$1.44} & {79.36$\pm$0.54} & {69.29$\pm$0.53}\\
    {FeSEM} & {65.65$\pm$1.28} & {76.46$\pm$0.63} & {68.49$\pm$1.00}\\
    {FedGroup} & {67.38$\pm$1.34} & {76.77$\pm$1.00} & {69.20$\pm$1.10} \\
    {CGPFL} & {71.19$\pm$0.93} & {79.42$\pm$0.46} & {70.33$\pm$0.46} \\
    {IFCA} & {73.06$\pm$0.91} & {80.54$\pm$0.74} & {72.02$\pm$1.13} \\
    {\textbf{FedAC}} & {\textbf{74.88$\pm$0.65}} & {\textbf{81.29$\pm$0.61}} & {\textbf{73.57$\pm$0.74}} \\
    \bottomrule
    \toprule
    \multirow{2}{*}{\textbf{Method}} &
    \multicolumn{3}{c}{CIFAR-100 Top-1 Test Accuracy (\%)} \\
    & {$\alpha=0.1$} & {$n=8$} & {$n=10$} \\
      \midrule
    {FedAvg} & {35.36$\pm$0.92} & {33.33$\pm$0.42} & {32.87$\pm$0.38} \\
    {FedPer} & {43.46$\pm$1.27} & {62.10$\pm$0.63} & {53.36$\pm$0.89}\\
    {FeSEM} & {31.03$\pm$0.80} & {52.15$\pm$2.45} & {48.15$\pm$2.74}\\
    {FedGroup} & {33.26$\pm$1.01} & {57.02$\pm$0.84} & {53.16$\pm$1.22} \\
    {CGPFL} & {41.38$\pm$0.69} & {60.26$\pm$0.94} & {50.49$\pm$1.02} \\
    {IFCA} & {38.61$\pm$0.77} & {60.18$\pm$0.76} & {57.68$\pm$0.53} \\
    {\textbf{LCFed}} & {\textbf{51.28$\pm$0.35}} & {\textbf{64.53$\pm$0.34}} & {\textbf{57.74$\pm$0.84}} \\
    \bottomrule
  \end{tabular}
  
\label{tab::overall}
\end{table}

Tabel~\ref{tab::overall} presents the converged accuracy and standard deviation on MNIST, CIFAR-10 and CIFAR-100 datasets. It is evident that LCFed outperforms other benchmarks in converged accuracy across diverse heterogeneous settings.
CFL methods perform better under pathological distribution because each device contains data for only a few labels. This leads to high data similarity within clusters (with shared labels) and minimal overlap between clusters, making knowledge sharing more effective within clusters.
IFCA outperforms FeSEM and FedGroup, deriving its advantage from clustering based on losses of cluster centers at the device side rather than model parameters. This approach optimizes clustering effectiveness but introduces substantial additional communication overhead, as discussed in Sec.~\ref{subsec::clus_cost}.
LCFed consistently maintains optimal performance across almost all settings, primarily due to its innovative integration of global knowledge into intra-cluster collaborative training. 
This allows local updates to learn knowledge suited to the data distribution within clusters while benefiting from low-dimensional global knowledge.
By fully leveraging knowledge across the system, LCFed maximizes training performance.

\begin{figure}[t]
\centering
\begin{subfigure}{0.23\textwidth}
\includegraphics[width=\textwidth]{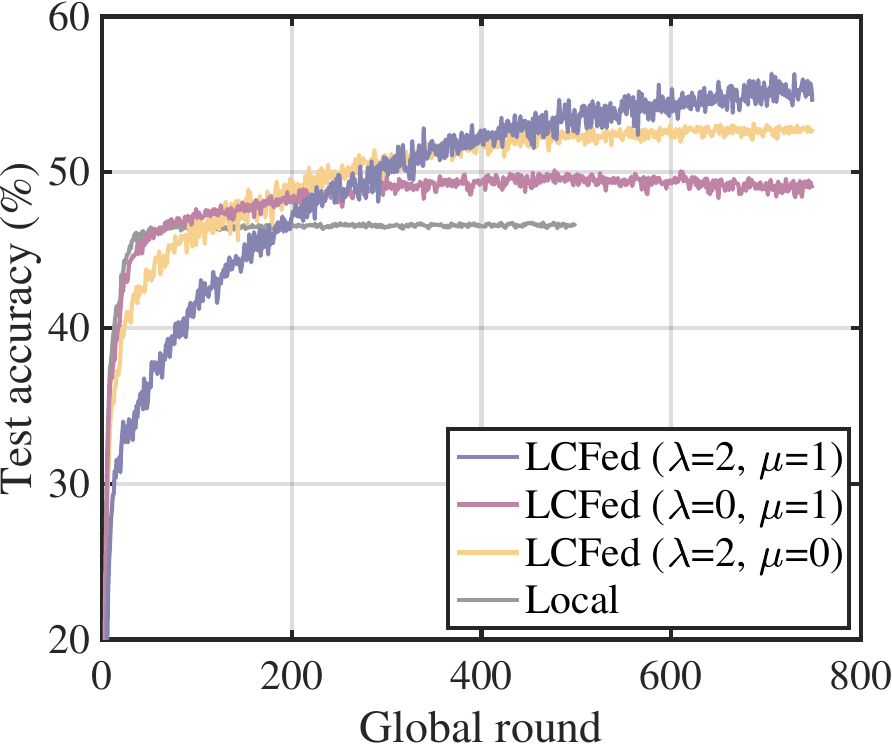}
    \caption{{CIFAR-100 ($\alpha=0.1$).}}
    \label{subfig:transbias}
\end{subfigure}
\hfill
\begin{subfigure}{0.23\textwidth}
\includegraphics[width=\textwidth]{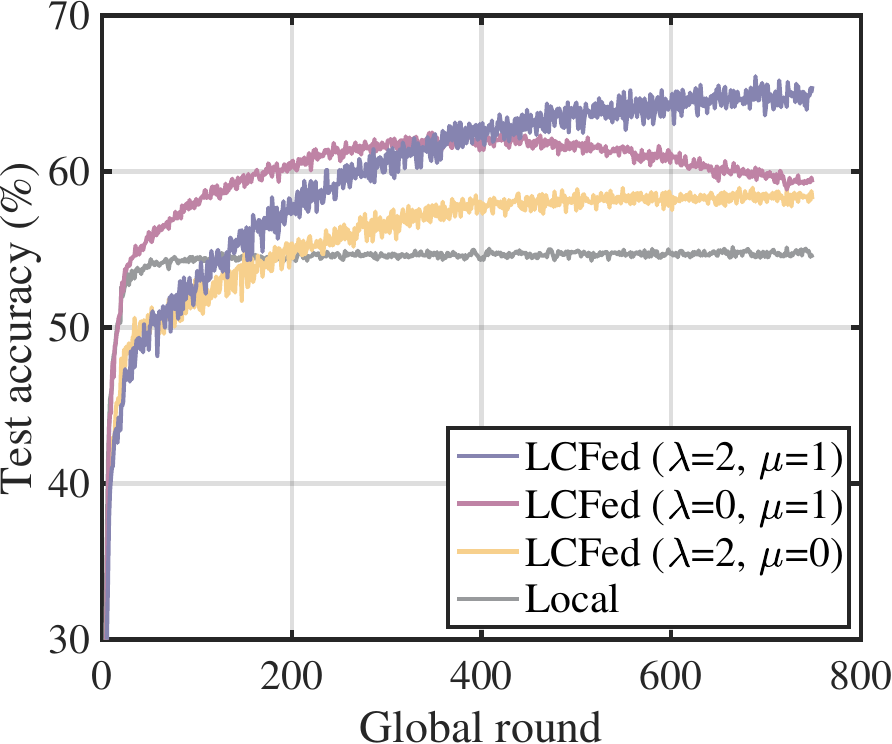}
    \caption{{CIFAR-100 ($n=8$).}}
    \label{subfig:transredun}
\end{subfigure}
\caption{The impact of regularization strengths $\mu$ and $\lambda$ on LCFed training performance.}
\label{fig:exp_abla}
\vspace{-3.5ex}
\end{figure}

Fig.~\ref{fig:exp_abla} presents the ablation experiments on global and intra-cluster knowledge in the design of LCFed's local update loss.
Under pathological distributions, $\mu=0$ leads to significant performance deterioration, indicating high intra-cluster similarity, thereby favoring intra-class knowledge.
It's worth noting that under both heterogeneous settings, optimal performance is achieved through the combined influence of global and intra-cluster knowledge. This validates the limitations of other CFL methods that confine knowledge sharing within clusters.

\subsection{Clustering Cost}
\label{subsec::clus_cost}

\begin{figure}[t]
\centering
\begin{subfigure}{0.235\textwidth}
\includegraphics[width=\textwidth]{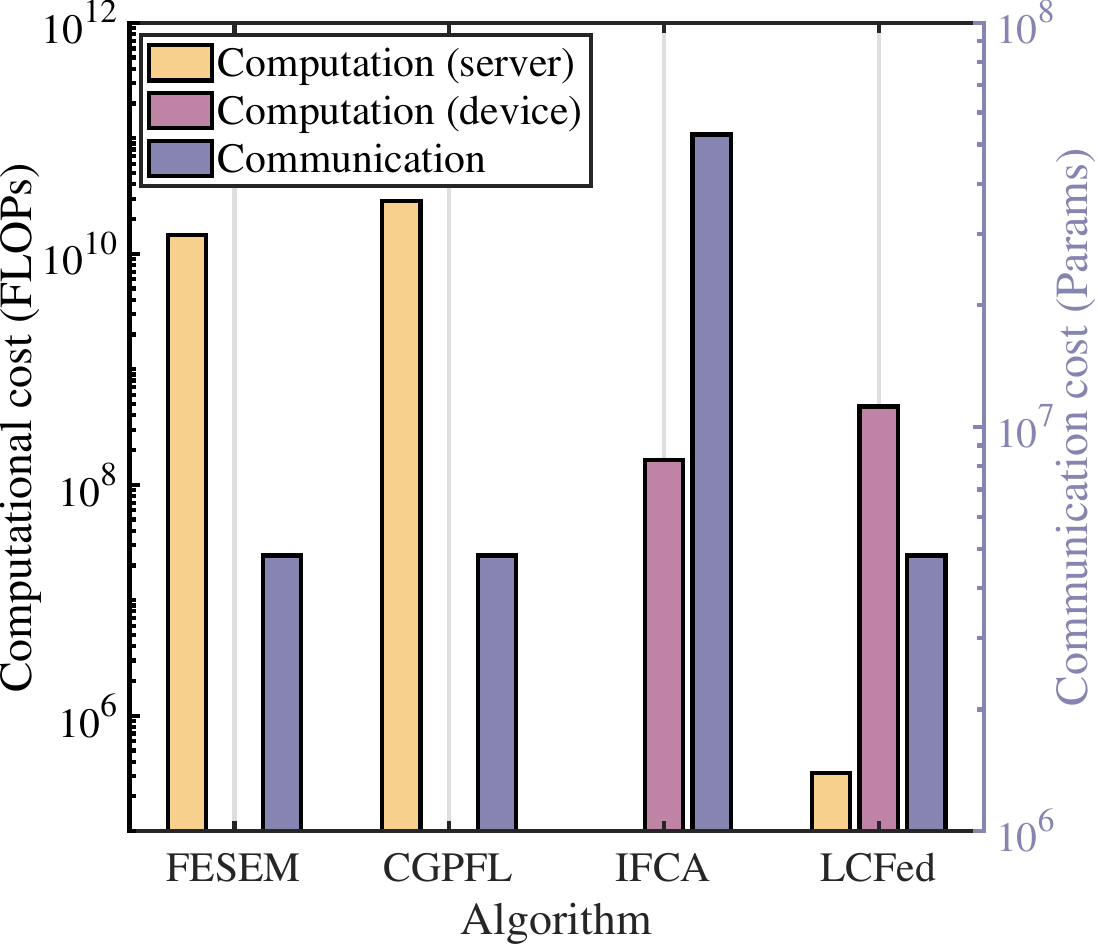}
    \caption{$m=100$, $K=10$.}
    \label{subfig:transbias}
\end{subfigure}
\hfill
\begin{subfigure}{0.235\textwidth}
\includegraphics[width=\textwidth]{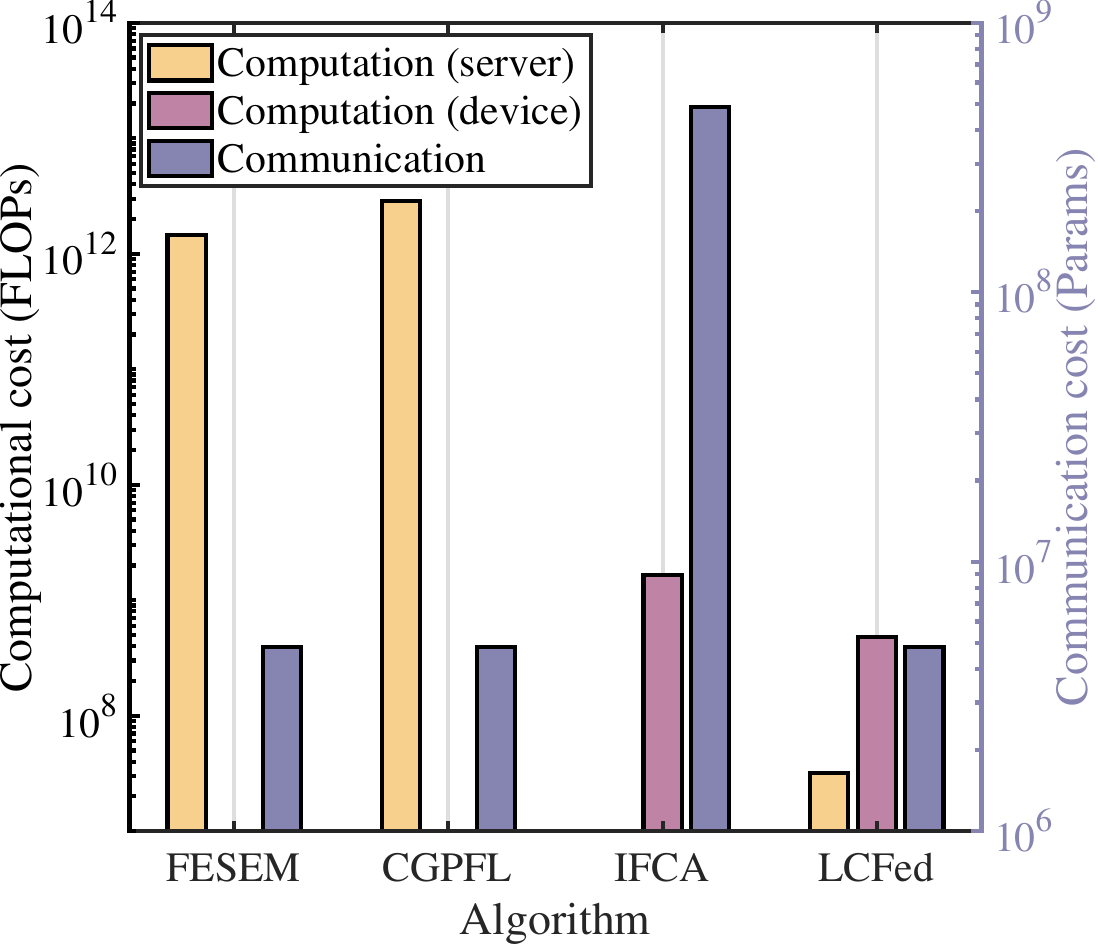}
    \caption{$m=1000$, $K=100$.}
    \label{subfig:transredun}
\end{subfigure}
\caption{Comparison of CFL algorithms in terms of clustering computation and communication costs across different scales.}
\label{fig:exp_cost}
\vspace{-3.5ex}
\end{figure}

Fig.~\ref{fig:exp_cost} shows the computational and communication costs per round of different online clustering methods in various scale scenarios.
FeSEM and CGPFL compare model parameters using negative L2 distance and cosine similarity as model similarities, respectively.
IFCA sends all cluster centers from server to edge devices, where the empirical loss is computed and compared for each cluster center.
Compared to FeSEM and CGPFL, which calculate similarity using full model parameters, LCFed reduces the server's computational overhead by over $9.048 \times 10^4$ times, lowering the parameter dimensionality of the LeNet-5 model from more than $4,800,000$ to $D=50$.
In comparison, IFCA's clustering does not require additional server-side computation, but transmitting $K$ cluster center models between from server to devices each round results in significant communication overhead.
Clearly, LCFed's low-rank model-based similarity measure shows optimal overall computational and communication efficiency. This advantage is even more pronounced in large-scale CFL scenarios, supporting the increasing prevalence of edge devices and IoT deployments in real-world applications.

\section{Conclusion}
\label{sec:conclusion}

This paper proposes LCFed, a novel and efficient CFL framework with two key innovations. First, LCFed partitions the model, utilizing specialized training and aggregation strategies for different sub-models, thereby effectively integrating global and intra-cluster knowledge to improve local training performance. Second, by leveraging a low-rank model-based similarity measure, LCFed significantly reduces computational overhead while maintaining robust clustering effectiveness. Extensive experiments demonstrate that LCFed surpasses state-of-the-art benchmarks in both test accuracy and computational efficiency.


\bibliography{refs}
\bibliographystyle{IEEEtran}

\end{document}